\documentclass[conference,a4paper]{IEEEtran}

\ifCLASSINFOpdf
  \usepackage[pdftex]{graphicx}
  \graphicspath{{../figures/}}
  \DeclareGraphicsExtensions{.pdf,.jpeg,.png}
\fi

\usepackage{graphicx}
\usepackage{amsmath,amssymb} 
\usepackage[colorlinks]{hyperref}
\usepackage{cleveref}
\usepackage{array}
\usepackage{caption}
\usepackage{subcaption}
\usepackage{multirow}

\usepackage{color}
\usepackage{array}

\newcolumntype{M}[1]{>{\centering\arraybackslash}m{#1}}

\begin{document}
\title{Image and Encoded Text Fusion for\\ Multi-Modal Classification}

\author{
\IEEEauthorblockN{I. Gallo, A. Calefati, S. Nawaz}

\IEEEauthorblockA{University of Insubria,\\
Dep. of Theoretical and Applied Science,\\
Varese, Italy\\
Email: ignazio.gallo@uninsubria.it}
\and
\IEEEauthorblockN{M. K. Janjua}

\IEEEauthorblockA{
  National University of \\ Sciences and Technology,\\
  Islamabad, Pakistan 
}}
\newpage
\onecolumn
\onecolumn{%
 \centering
 \LARGE DICTA Publication License for PID \# 1222850\\[1.5em]
  \large \textbf{Paper Title:} Image and Encoded Text Fusion for Multi-Modal Classification\\[1em]
 \large \textbf{Author:} I. Gallo, A. Calefati, S. Nawaz, M.K. Janjua\\[1em]}
\large
\section{License}
Copyright 2018 IEEE. Published in the Digital Image Computing: Techniques and Applications, 2018 (DICTA 2018), 10-13 December 2018 in Canberra, Australia. Personal use of this material is permitted. However, permission to reprint/republish this material for advertising or promotional purposes or for creating new collective works for resale or redistribution to servers or lists, or to reuse any copyrighted component of this work in other works, must be obtained from the IEEE. Contact: Manager, Copyrights and Permissions / IEEE Service Center / 445 Hoes Lane / P.O. Box 1331 / Piscataway, NJ 08855-1331, USA. Telephone: + Intl. 908-562-3966
\twocolumn
\maketitle
\begin{abstract}
Multi-modal approaches employ data from multiple input streams such as textual and visual domains. Deep neural networks have been successfully employed for these approaches. In this paper, we present a novel multi-modal approach that fuses images and text descriptions to improve multi-modal classification performance in real-world scenarios. The proposed approach embeds an encoded text onto an image to obtain an information enriched image. To learn feature representations of resulting images, standard Convolutional Neural Networks (CNNs) is employed for classification task. We demonstrate how a CNN based pipeline can be used to learn representations of the novel fusion approach. We compare our approach with individual sources on two large scale multi-modal classification datasets while obtaining encouraging results. Furthermore, we evaluate our approach against two famous multi-modal strategies namely early fusion and late fusion. 
\end{abstract}

\begin{figure}[ht!]
  \centering 
  \begin{subfigure}[b]{0.40\linewidth}
    \centering\includegraphics[width=\textwidth]{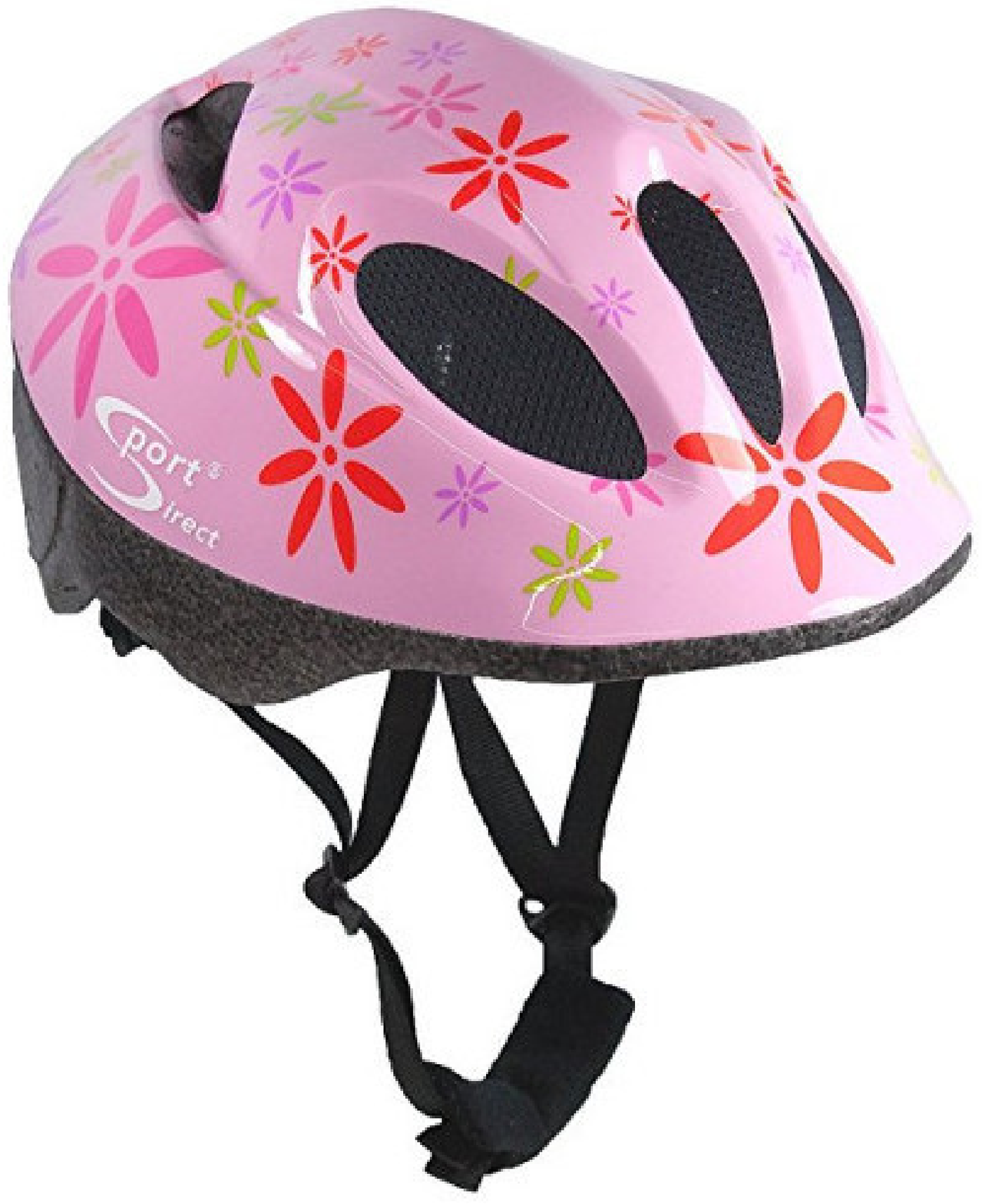}
    \caption{Useful accessory for those who ride a \textbf{bike}. Size 46-52. \label{subfig:helmet}}
  \end{subfigure}~~~
  \begin{subfigure}[b]{0.40\linewidth}
    \centering\includegraphics[width=\textwidth]{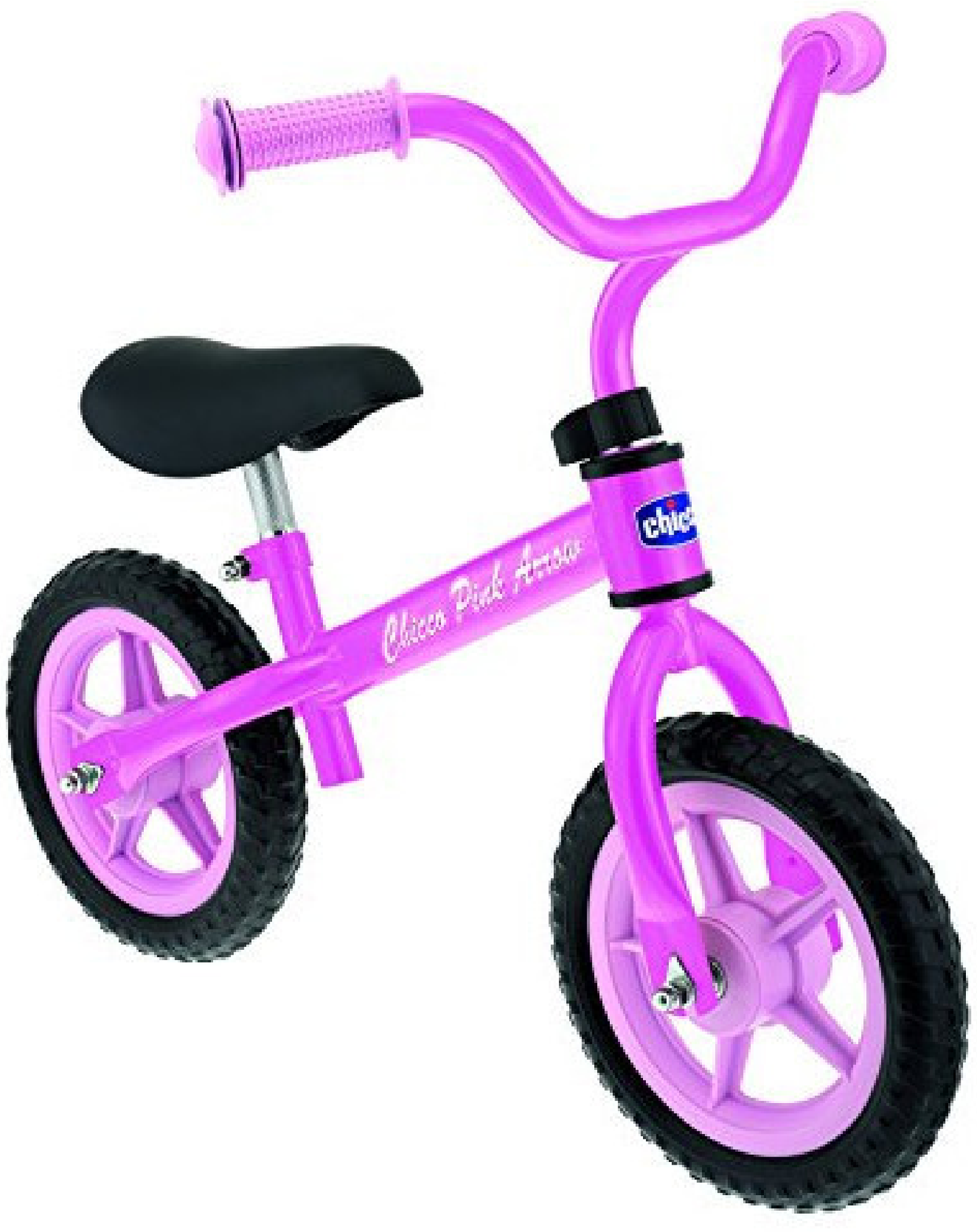}
    \caption{The First \textbf{Bike} Pink Arrow dedicated to little girls.  \label{subfig:bike}}
  \end{subfigure}\\
  \begin{subfigure}[b]{0.45\linewidth}
    \centering\includegraphics[width=\textwidth]{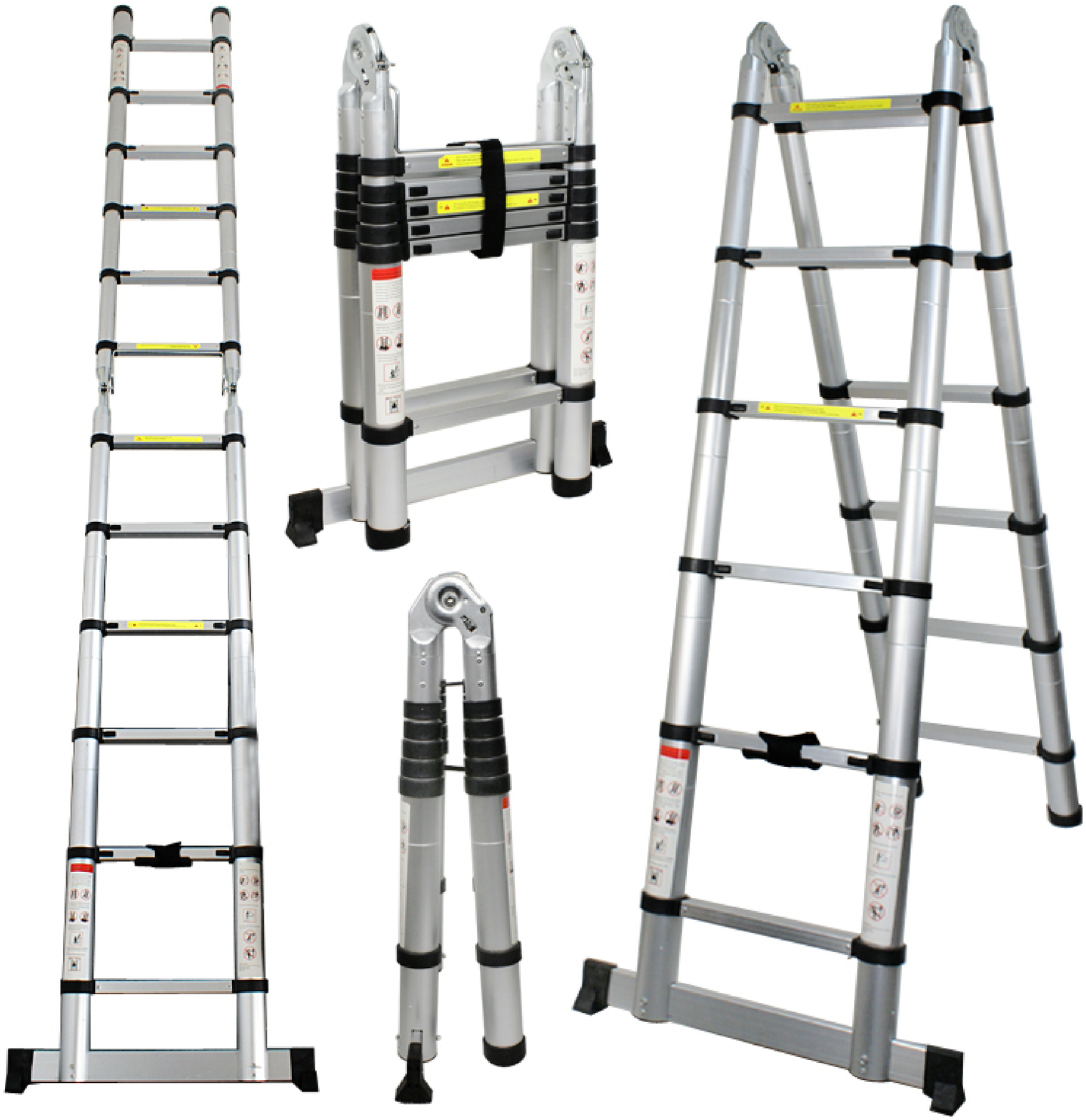}
    \caption{Telescopic \textbf{ladder} to partial or total opening. Ideal for any external intervention.\label{subfig:ladder}}
  \end{subfigure}~~~%
  \begin{subfigure}[b]{0.45\linewidth}
    \centering\includegraphics[width=\textwidth]{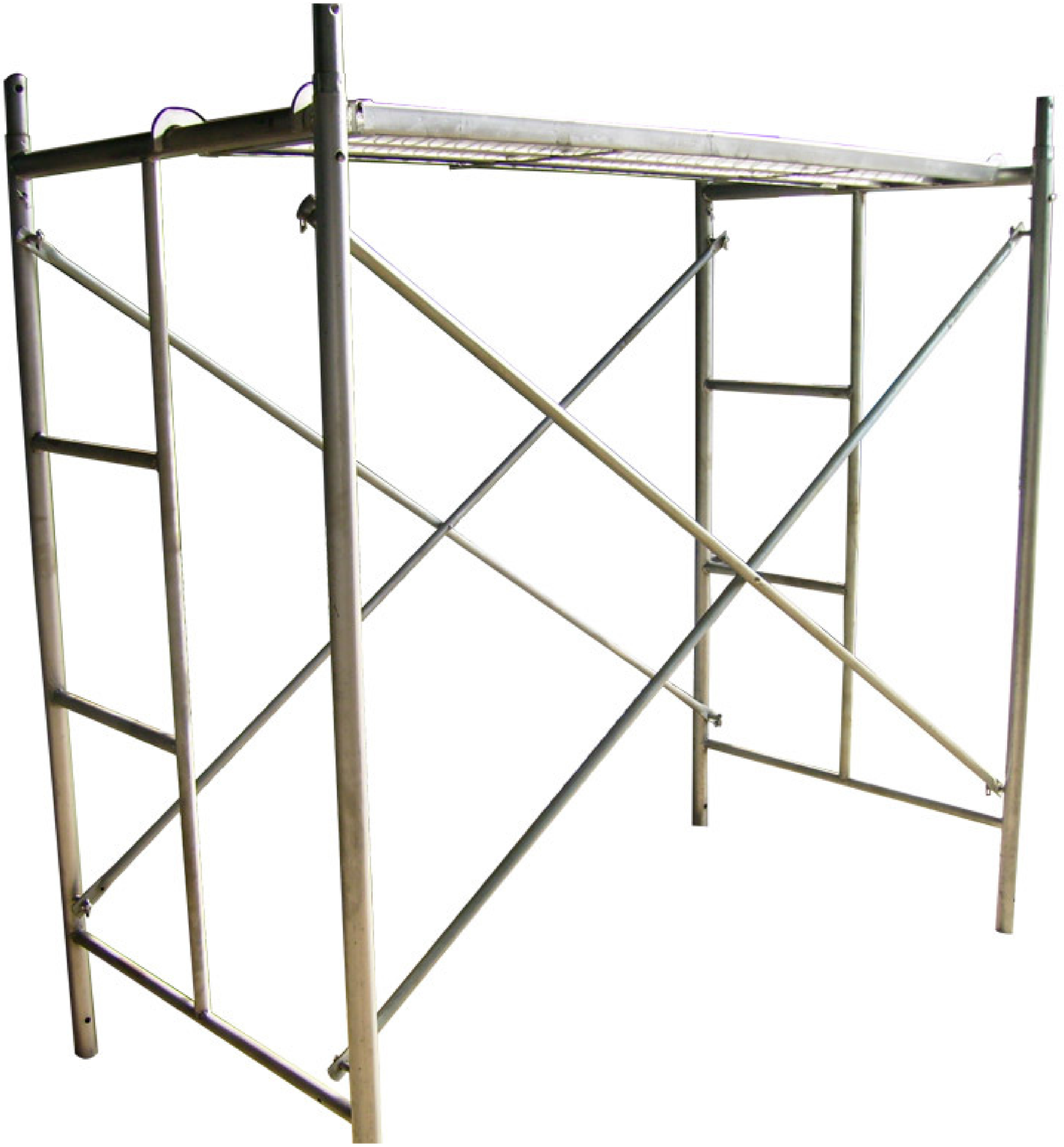}
    \caption{Custom multifunction dynamic construction \textbf{scaffolding}, simple for decoration.\label{subfig:scaffolding}}
  \end{subfigure}
  \caption{In the top row, an example of ambiguous text descriptions that can be disambiguated with the analysis of the accompanying images.
  In the bottom row, an examples of ambiguous images that can be disambiguated with the analysis of the associated text descriptions.
  }
  \label{fig:the-problem}
\end{figure}

\section{Introduction}
With the rapid rise of e-commerce, the web has increasingly become multi-modal, making the question of multi-modal strategy ever more important.
However, modalities  in multi-modal approach come from different input sources (text/image~\cite{gallo2017multimodal,kiela:2014:Learning,kiela2018efficient} , audio/video~\cite{ngiam2011multimodal} etc.) and are often characterized by distinct statistical properties, making it difficult to create a joint representation that uniquely captures the ``concept'' in the real-world applications.
For example, Figure~\ref{fig:the-problem} shows two adverts typically available on e-commerce website, where two objects have seemingly similar text descriptions in first row but seemingly different images. 
On the other hand we have two different text descriptions but similar images in the second row.
This leads us to create a joint representation of an image and text description for this classification problem.
Multi-modal strategy can exploit such scenario to remove ambiguity and improve classification performance.

The use of multi-modal approach based on image and text features is extensively employed on a variety of tasks including modeling semantic relatedness, compositionality,
classification and retrieval~\cite{guillaumin2010multimodal,kiela:2014:Learning,leong2011going,feng2010visual,kiela2018efficient,wang2016learning}.
Typically, in multi-modal approach, image features are extracted using CNNs. 
Whereas, to generate text features, Bag-of-Words models or Log-linear Skip-gram Models~\cite{mikolov2013efficient} are commonly employed. 
This represents a challenge to find relationships between features of multiple modalities along with representation, translation, alignment, and co-learning as stated in~\cite{baltruvsaitis2018multimodal}.

With this work, we present a novel strategy which combines a text encoding schema to fuse text features and image in an unified information enriched image.
We merge both text encoding and image into a single source so that it can be used with a CNN.
We demonstrate that by adding encoded text information in an image, multi-modal classification results can be improved compared to the best one obtained on a uni-modality (image/text).

Intuitively superimposing text descriptions onto images may not sound motivating due to several reasons. Since the idea is overlaying the encoded text description onto an image, it might affect the image perception in general. However, this is not true, the main strength of the approach is that embedded text can be overlaid onto the image with fixed width regardless of the size of text description. We experiment with different embedding sizes to verify that image perception is not affected. 
Figure~\ref{fig:text-vs-multimodal} plots different embedding sizes to explain the network behavior under such scenario. 

Main contributions of our paper are listed below: 
\begin{itemize}
  \item We present a novel data fusion mechanism based on encoded text description and associated image for multi-modal classification.
  \item We show that fused data is classified with a standard CNN based architectures, typically employed in image classification.
  \item We evaluate the fused multi-modal approach on two large scale datasets to show the effectiveness of our approach. 
\end{itemize}

\begin{figure*}[t]
  \centering 
  \includegraphics[width=0.9\textwidth]{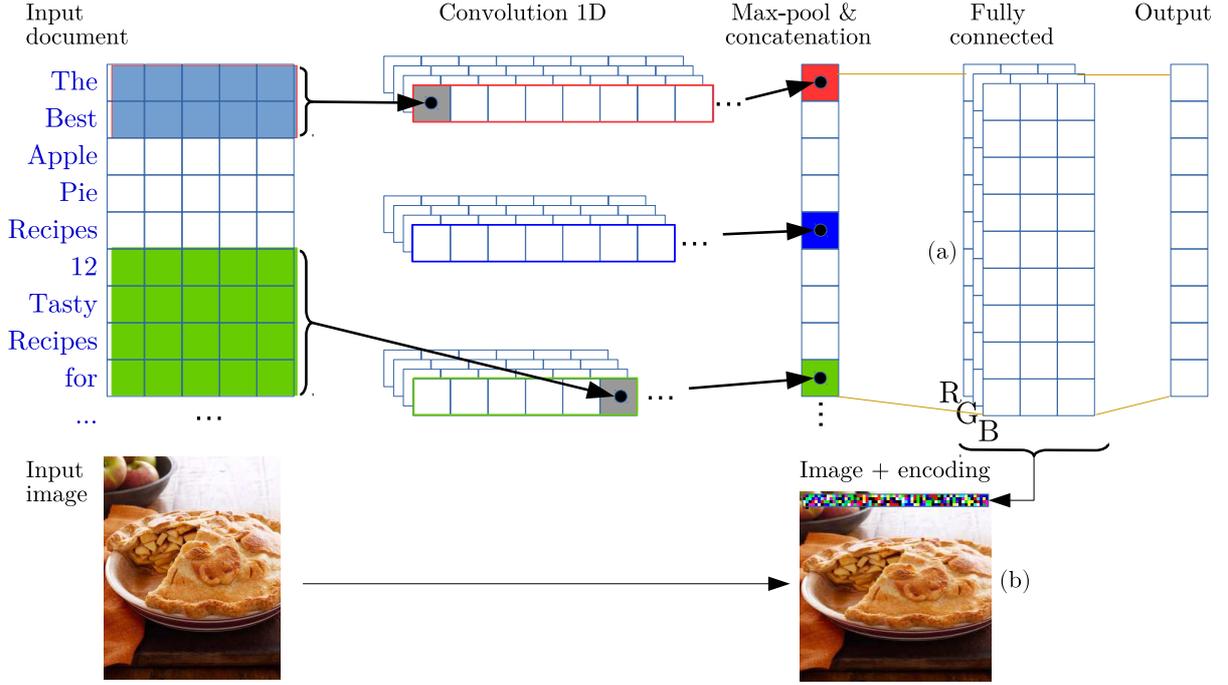}
  \caption{The proposed text and image fusion model for deep multi-modal classification. 
  The text encoded (a) is passed to the output layer.
  After the training step only the text features (a) are extracted and then drawn over the original image in order to generate a new multimodal dataset.
  }
  \label{fig:proposed-model}
\end{figure*}

\section{Related Work}

There are two general multi-modal fusion strategies to fuse text and images features, namely early fusion and late fusion~\cite{atrey2010multimodal,baltruvsaitis2018multimodal}, each one having
its advantages and disadvantages.

Early fusion is an initial attempt by the researchers towards multi-modal representation learning. 
Early fusion methods concatenates text and image features into a single vector which is used as input pattern for the final classifier. 
The technique is employed for various tasks~\cite{Bruni:2011:Distributional,kiela:2014:Learning,kiela2018efficient}.
The main benefit of early fusion is that it can learn to exploit the correlation and interactions between low level features of each modality.

In contrast, late fusion~\cite{poria2015deep} uses decision values from each modal and fuses them using a fusion mechanism. 
Multiple works ~\cite{shutova2016black,evangelopoulos2013multimodal} employ different fusion mechanisms such as averaging, voting schemes, variance etc.
The work in~\cite{gallo2017multimodal} showcased a comparative study of early and late fusion multi-modal methods.   
Late fusion produced better performance compared to early fusion method, however, it comes with the price of an increased learning effort.  
In addition, a strategy must be introduced to assign a weight to each classifier employed.
This presents another challenge in late fusion strategy.
Our method is inspired from early fusion~\cite{kiela:2014:Learning}, however, taking advantage of the idea of our previous work~\cite{gallo2017semantic} we concatenate encoded text features into an image to obtain an information enriched image.
In this work, we encode text features onto an image with an encoding schema similar to the one proposed in~\cite{gallo2017semantic}.
The main difference lies in the type of embedding used: our previous work~\cite{gallo2017semantic} used the encoding extracted from Word2Vec and therefore we obtained a numeric vector for each word in a text document, while in this work we extract text features from a CNN network for text classification, trained using all the words available in a description.
In the next sections, the encoding technique used to graphically represent the text above the image, will be summarized.

Multi-modal fusion methods are successfully employed to other modalities, e.g. video and audio ~\cite{ngiam2011multimodal,nagrani2018seeing}.
Other interesting examples of multi-modal approaches that make use of deep networks include restricted Boltzmann machines~\cite{srivastava2012multimodal}, auto-encoders~\cite{wu2013online}.

\begin{table}
\begin{center}
\caption{Network configuration summary. $k$, $s$ and $p$ stand for kernel size, stride and padding size
respectively. In the convolution layer, we use 128 filters for each of the following sizes 3,4,5 (the first one is showed below).
The embedding size is $w=128$.
}
\label{tab:network-info}
\begin{tabular}{lccccc}
\hline\noalign{\smallskip}
Type & Configuration \\
\noalign{\smallskip}
\hline
\noalign{\smallskip}
Output             & num. classes        \\
Fully Connected    & $H_t$(encoded-text-h) $\times$ $W_t$(encoded-text-w) $\times 3$     \\  
MaxPool-1D 	   & $h$:$S-k+1$, $k$:1, $s$:1, $p$:1       \\ 
Convolutional-1D   & $w$(embedding-size), $k$:$3\times w$, $s$:1, $p$:1     \\ 
Input  	 	   & $S$: 100 words (sequence length)        \\
\hline
\end{tabular}
\end{center}
\end{table}

\section{The Proposed Approach}
\label{proposed-approach}
%
In this work we take a cue from our previous work~\cite{gallo2017semantic} to transform a text document onto an image to be classified with a CNN.
However, instead of using numeric values from Word2Vec model to represent a text document, we are using a new approach involving a CNN trained for text classification.

First, we transform the text document into a visual representation to construct an information enriched image containing text features and image.
Finally, we solve the multi-modal problem using this image to train a CNN generally used for image classification.

We use a variant of the CNN model proposed by Kim~\cite{kim2014convolutional} for text document classification.
The input layer is a text document followed by a convolution layer with multiple filters, then a max-pooling layer followed by a fully connected layer, and finally a softmax classifier. 
The network configuration summary is show in Table~\ref{tab:network-info}. 
Text features are extracted from the fully connected layer (Figure~\ref{fig:proposed-model}a) and transformed into an RGB encoding so that it can be overlaid onto an image associated with the text document.
Figure~\ref{fig:proposed-model} shows architectural representation of the model used to encode the text dataset into an image dataset (Figure~\ref{fig:proposed-model}b) to obtain a multi-modal dataset.
In the second step, resulting images are fed to any baseline CNN for classification.

The major advantage of our method is that we can cast a uni-modal into a multi-modal CNN without the need of adapting the model itself.
This approach is suitable to be adopted in multi-modal methods because a CNN architecture can extract information from both the encoded text and the related image.

\subsection{Encoding Scheme}
\label{sec:proposed-encoding-technique}
We exploit the CNN model proposed by Kim~\cite{kim2014convolutional} which performs text to visual features transformation within a single step.
Figure~\ref{fig:proposed-model} summarizes the encoding system used in this work, where a reshape was applied to the fully connected layer showed in Figure~\ref{fig:proposed-model}a to transform an array into an image representing the encoded text to be superimposed on the original image.

\begin{figure*}[t]
  \centering 
  \begin{tabular}{ll|ll}
  \multicolumn{2}{c}{UMPC Food-101} & \multicolumn{2}{c}{Ferramenta} \\ 
  \noalign{\smallskip}
  \hline
  \noalign{\smallskip}
  \includegraphics[width=0.215\textwidth]{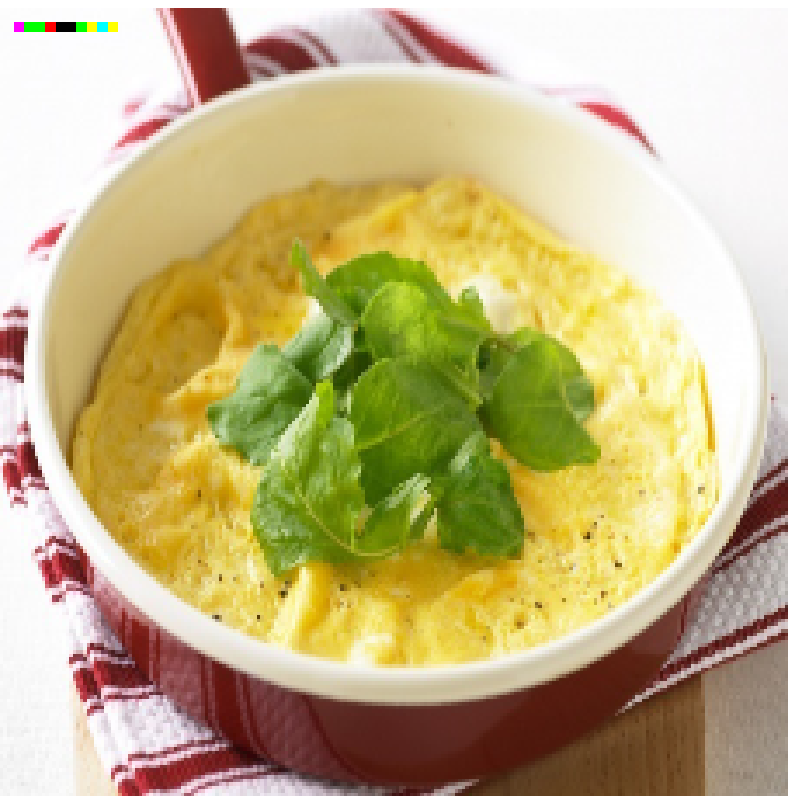} & 
  \includegraphics[width=0.215\textwidth]{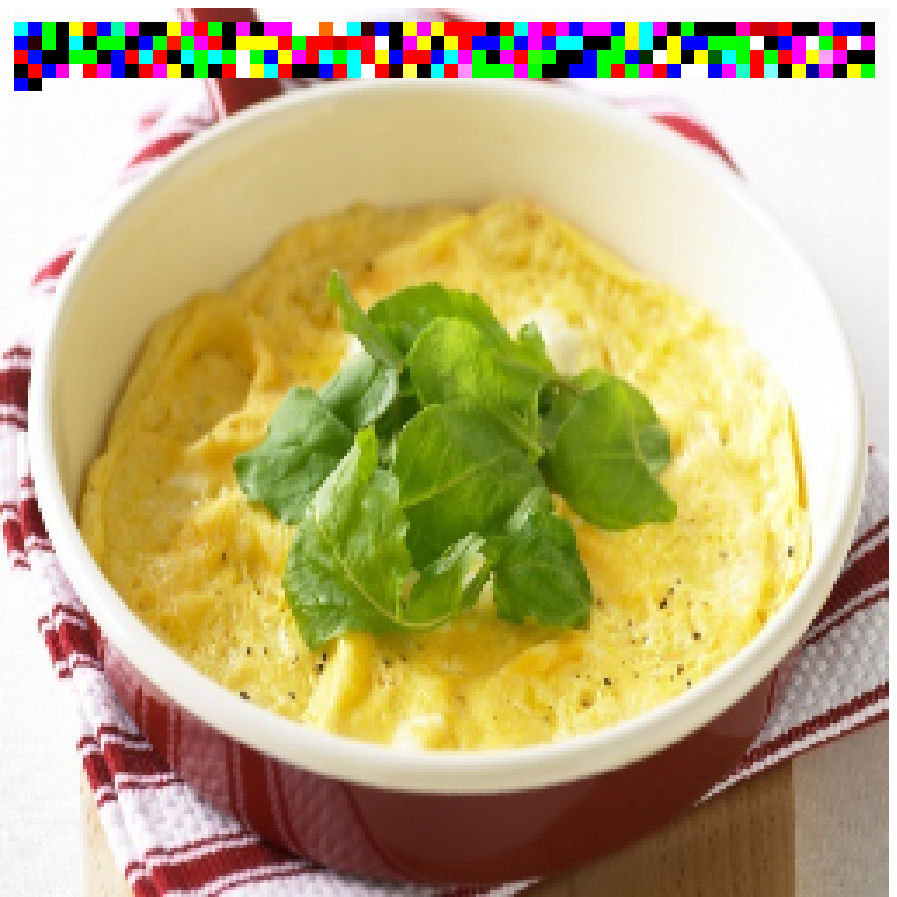} & 
  \includegraphics[width=0.215\textwidth]{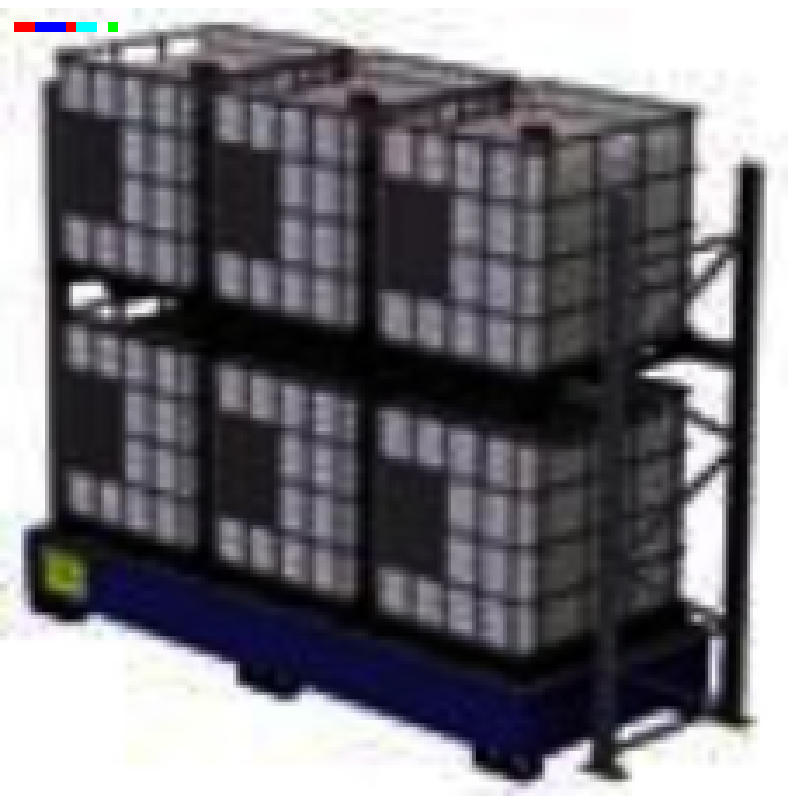} & 
  \includegraphics[width=0.215\textwidth]{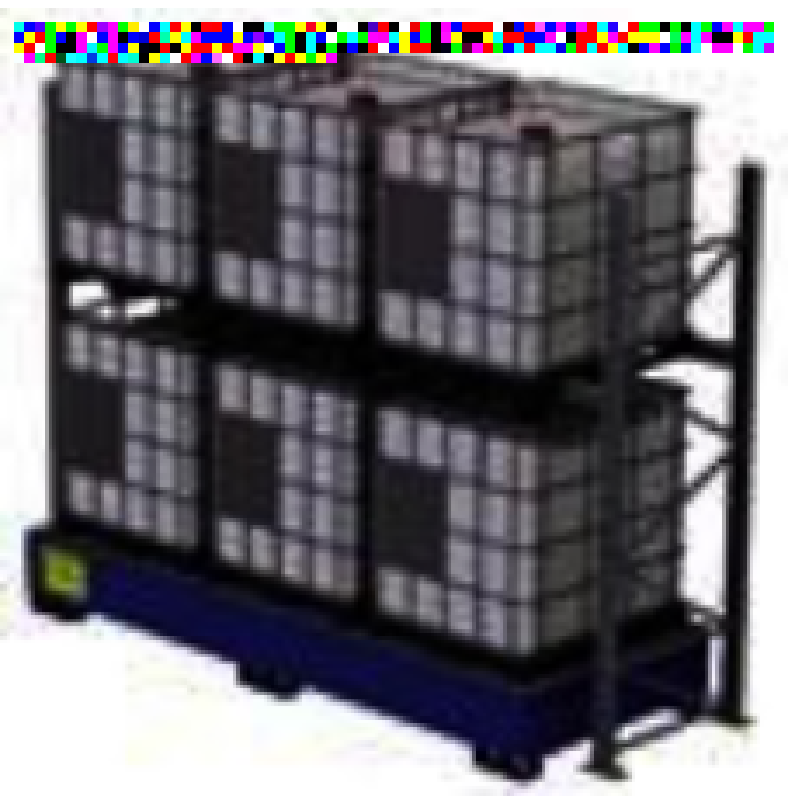}
  \end{tabular}
  \caption{Two encoding examples taken from two datasets.
   Images on the left column show an encoding of $10$ superpixel length while on the right column we have an encoding of length equal to $250$ superpixel.
   All images are $227\times 227$ in size having an encoding superpixel equals to $3\times 3$ pixels for Ferramenta dataset and $4\times 4$ for Food-101 dataset. }
  \label{fig:encoding-examples}
\end{figure*}

Features are extracted from the trained CNN model and transformed into a visual representation of the document.
In practice, we used feature vectors showed in Figure~\ref{fig:proposed-model}a, having a size $L=3\cdot w\cdot h$ that is a multiple of 3 in order to be transformed into a color image.
We used the same concept of \textit{superpixel} used in~\cite{gallo2017semantic} to represent a sequence of three values $\in L$ as an area with a uniform color of $P\times P$ dimension.
In this way textual features are represented as a sequence of superpixel, drawn from left to right and from top to bottom, starting from a certain position of the scaled image (see some examples of the final multi-modal image in Figure~\ref{fig:encoding-examples} and Figure~\ref{fig:very-similar-encoding}).
Finally, we encode an entire text document within the image plane and then the next multi-modal CNN model can work simultaneously on both modalities.

This approach has an advantage to the work in~\cite{gallo2017semantic}, in fact in our work it is possible to encode long text documents because we encode the entire document in the same image area having fixed size equals to $w\times h\times 3$.

\section{Datasets}

In multi-modal dataset, modalities are obtained from different input sources.
Datasets used in this work consist of images and accompanying text descriptions.
We select Ferramenta~\cite{gallo2017multimodal} multi-modal dataset that are created from e-commerce website.
Furthermore, we select UMPC Food-101~\cite{wang2015recipe} multi-modal dataset to show the applicability of our approach to other domains.
Table~\ref{tab:dataset-info} shows information on these datasets.
The first column shows the number of class labels available in datasets.
While second and third columns show split on train and test sets.
The last column indicates the language of text description available for these datasets.  
Table~\ref{tab:img-text} shows image and associated text description randomly selected from each multi-modal dataset.

\begin{table}[t!]
\begin{center}
\caption{Information on multi-modal datasets used in this work. A multi-modal dataset consists of an image and accompanying text description. The last column indicates the text description language.}
\label{tab:dataset-info}
\begin{tabular}{lccccc}
\hline\noalign{\smallskip}
Dataset & \#Cls & Train  & Test  &  Lang. \\
\noalign{\smallskip}
\hline
\noalign{\smallskip}
Ferramenta   	& 52		& 66,141	& 21,869	& IT \\
Food-101 			& 101		& 67,988 	& 22,716	& EN \\
\hline
\end{tabular}
\end{center}
\end{table}

\begin{table*}[t!]
\begin{center}
\caption{An image and an associated text description randomly taken from each multi-modal dataset. Text descriptions in Ferramenta multi-modal dataset is translated from Italian to English for readers. UMPC Food-101 multi-modal dataset contains long text descriptions for food recipes however, we include a short text description.}
\label{tab:img-text}
\begin{tabular}{l m{2.6cm} m{6.0cm}}
\noalign{\smallskip}
\hline
\noalign{\smallskip}

Dataset & Image & Text Description  \\ 
\noalign{\smallskip}
\hline
\noalign{\smallskip}
Ferramenta~\cite{gallo2017multimodal}	&	\includegraphics[width=0.14\textwidth]{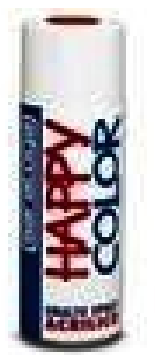}	& saratoga chestnut brown spray paint 400 ml happy color, quick-drying bright spray enamel for interiors and exteriors for applications on furniture chairs doors frames ornaments and all surfaces in wood metal ceramic glass plaster and masonite.   \\  \\
UPMC Food-101~\cite{wang2015recipe}  &	\includegraphics[width=0.14\textwidth]{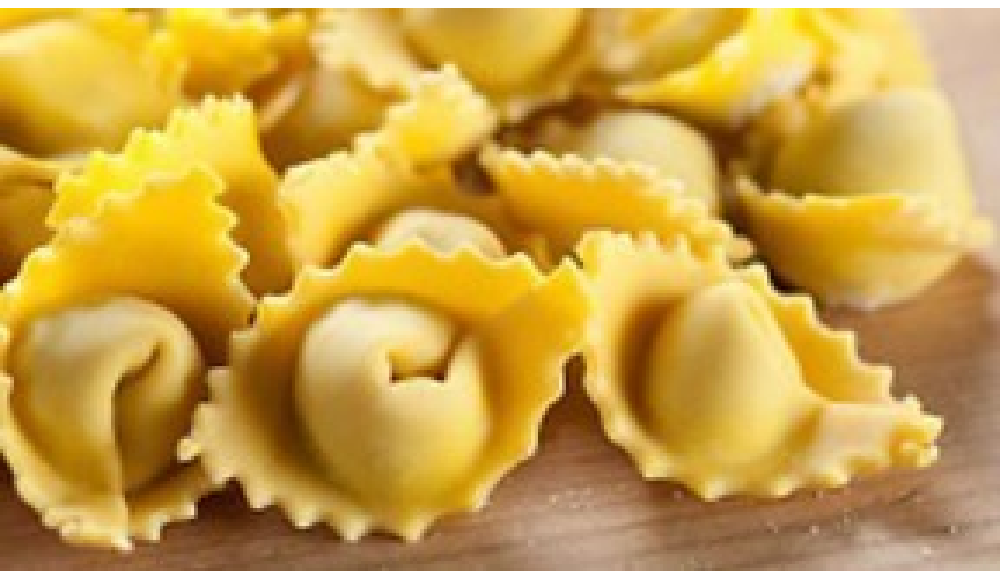}	&	Robiola-Cheese-Filled Ravioli Recipe  Pasta Recipes  ...\\ \\ \hline
    
\end{tabular}
\end{center}
\end{table*}

\begin{table}
\begin{center}
\caption{Classification results comparison on only-text, only-image and fused images. There are two baseline models for images and fused-images, while we use only one baseline for text-only scores.}
\label{tab:main-results}
\begin{tabular}{lccccccc}
\hline\noalign{\smallskip}
Dataset & Text & \multicolumn{2}{c}{Image}  & \multicolumn{2}{c}{Fusion}  \\ &  & AlexNet & GoogleNet &  AlexNet & GoogleNet\\ 
\noalign{\smallskip}
\hline
\noalign{\smallskip}
Ferramenta  	 	&  92.09  & 92.36		& 92.47 & 95.15  &  95.45 \\
Food-101  	 	&  79.78  & 42.01		& 55.65 & 82.90  &  83.37 \\
\hline
\end{tabular}
\end{center}
\end{table}

Ferramenta multi-modal dataset~\cite{gallo2017semantic} consists of $88,010$ adverts split in $66,141$ adverts for train set and $21,869$ adverts for test set, belonging to $52$ classes.
Ferramenta dataset provides a text and representative image for each commercial advertisement.
It is interesting to note that text descriptions in this dataset are in Italian Language. 

The second dataset that we use in our experiments is named UPMC Food-101 multi-modal dataset ~\cite{wang2015recipe}, containing about $100,000$ items of food recipes classified in $101$ classes. 
This dataset is collected from the web and each item consists of an image and the HTML webpage on which it was found. 
We have extracted the title from HTML document to use it in lieu of text description. 
Classes in the dataset are the $101$ most popular categories from the food picture sharing website\footnote{www.foodspotting.com}.

\section{Experiments}

\subsection{Preprocessing}

The proposed multi-modal approach transforms text descriptions and embeds them onto associated images to obtain information enriched images.
An example of information enriched image is shown in Figure~\ref{fig:encoding-examples}.
In this work, the transformed text description is embedded into a RGB image with an image size of $227\times227$ for UMPC Food-101 and Ferramenta multi-modal datasets.

\begin{figure}
  \centering 
  \includegraphics[width=1.0\columnwidth]{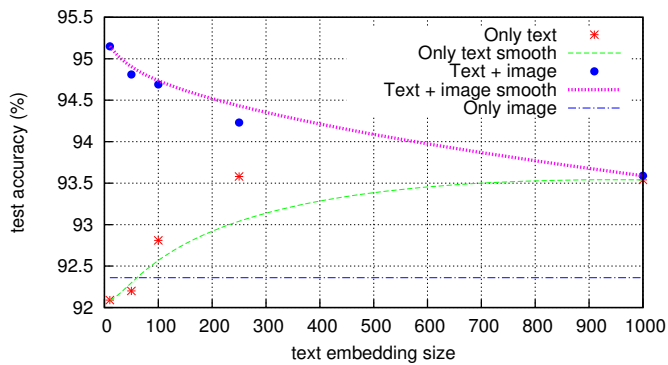}
  \caption{Comparison between the CNN that uses only text documents and only images with the CNN that uses fusion of image and encoded text, as the dimension of the text embedding varies. 
  In this experiment the Ferramenta multi-modal dataset is used.
  }
  \label{fig:text-vs-multimodal}
\end{figure}

\subsection{Detailed CNN settings}

We use a standard AlexNet~\cite{krizhevsky2012imagenet} and GoogleNet~\cite{Szegedy_2015_CVPR} on the Deep Learning GPU Training System (DIGITS) with default configuration. 
For fair comparison, we use same CNN settings for experiments using only images and fused images.
We use standard CNN hyperparameters. The initial learning rate is set to 0.01 along with Stochastic Gradient Descent (SGD) as optimizer. The network is trained for a total of $60$ epochs and/or till no further improvement is noticed to avoid over fitting. 
In our experiments, accuracy is used to measure classification performances.
The aim of the experiment is to show that by adding encoded text information in images it is possible to obtain better classification results compared to the best one obtained using a uni-modal (Text/Image).
We conducted following experiments with this aim in mind: 
(1) classification with CNN using only images, 
(2) classification with CNN using only text descriptions,
(3) classification with CNN using fused images,
(4) comparison with early and late fusion strategies.

\begin{table}[t!]
\begin{center}
\caption{Comparison of our approach with early and late fusion strategies. 
The results on the Ferramenta dataset are extracted from paper~\cite{gallo2017multimodal}}
\label{tab:early-late-results}
\begin{tabular}{lccccccc}
\hline\noalign{\smallskip}
Dataset & Early F. & Late F. & Proposed  \\ 
\noalign{\smallskip}
\hline
\noalign{\smallskip}
Ferramenta  	 	&  89.53  & 94.42		& \textbf{95.15} \\
Food-101  	 	&  60.83  & 34.43		& \textbf{82.90}  \\
\hline
\end{tabular}
\end{center}
\end{table}

\begin{figure*}
  \centering 
  \begin{tabular}{m{4cm}m{2cm}m{4cm}m{2cm}}
  \hline
  \multicolumn{2}{c|}{Ferramenta} & \multicolumn{2}{c}{UMPC Food-101} \\ \hline
  \fbox{\includegraphics[width=0.215\textwidth]{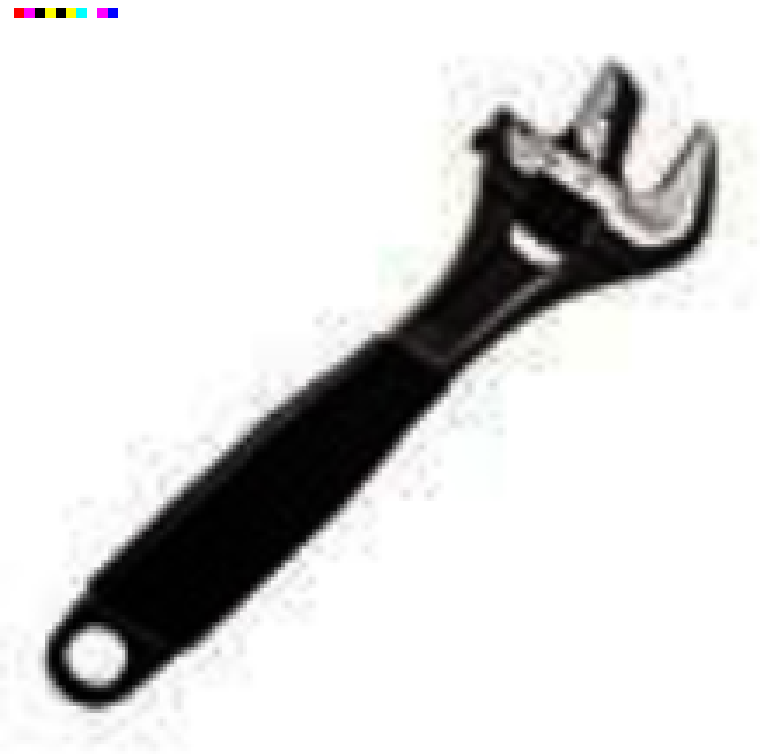}} & 
  bahco 9070p chiave inglese regolabile ergonomica 15 3 cm 6 pollici a becco reversibile colore nero & 
  \fbox{\includegraphics[width=0.215\textwidth]{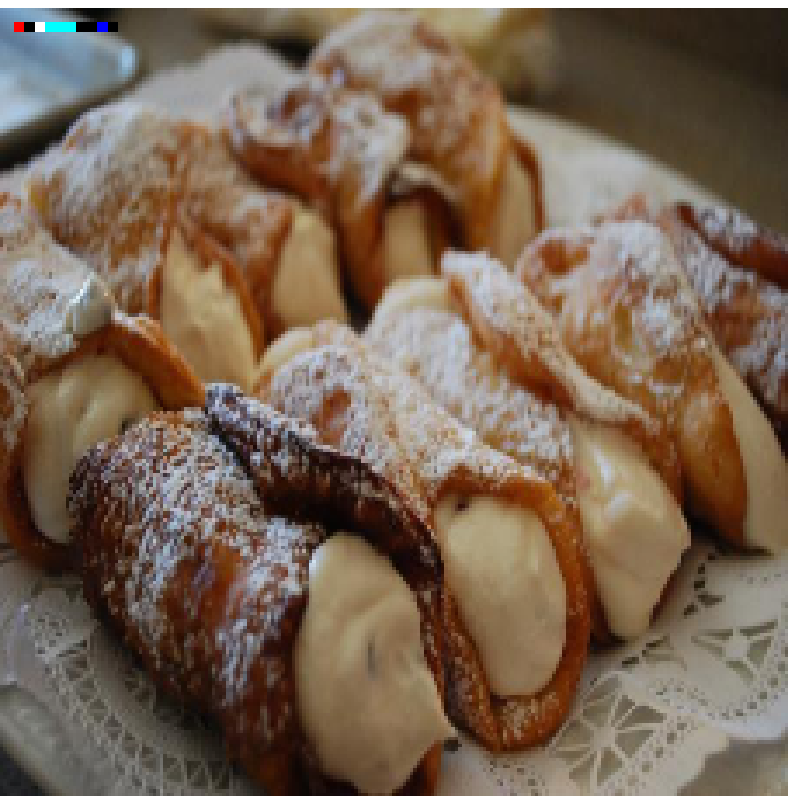}} & Cannoli Recipe - Food.com \\
  \fbox{\includegraphics[width=0.215\textwidth]{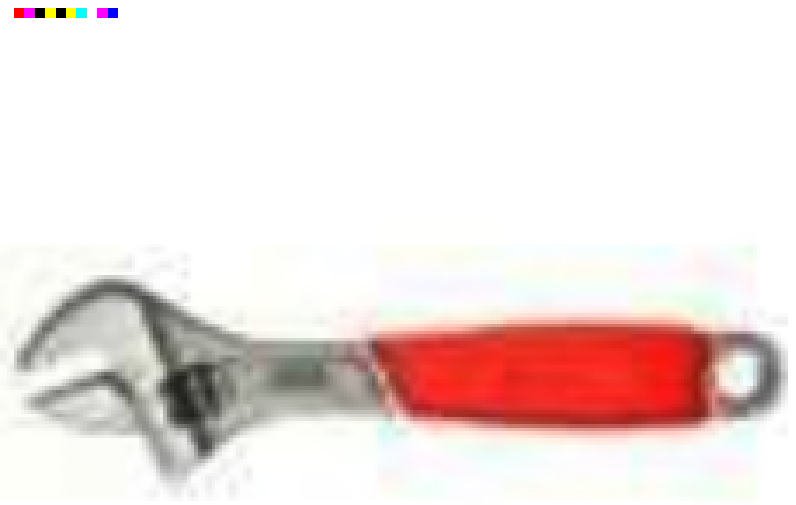}} & 
  connex cox550110 chiave inglese regolabile 25 4 cm  & 
  \fbox{\includegraphics[width=0.215\textwidth]{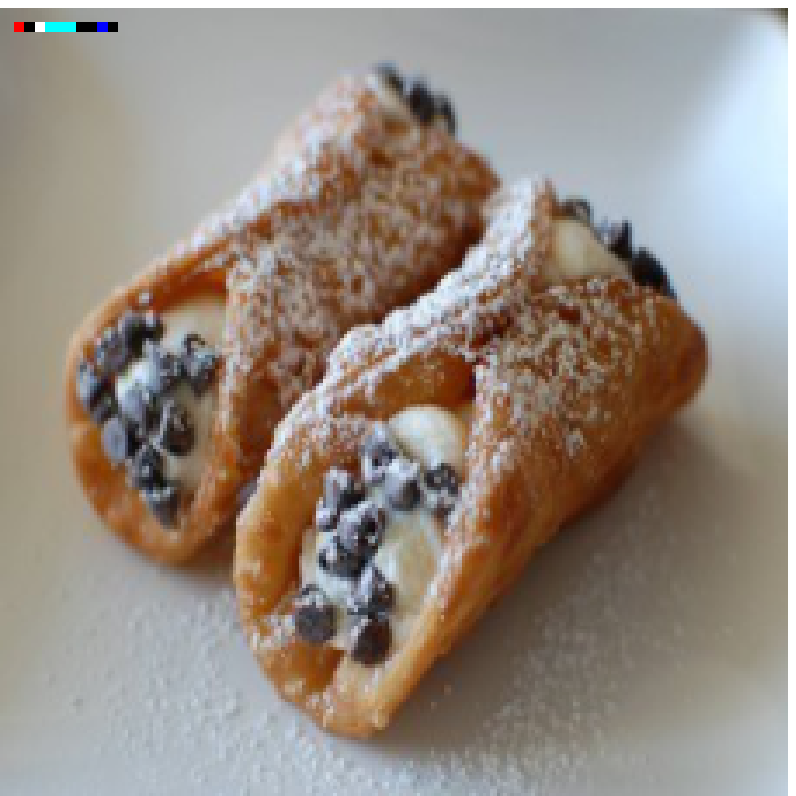}} & homemade cannoli filling The 350 Degree Oven \\
  \fbox{\includegraphics[width=0.215\textwidth]{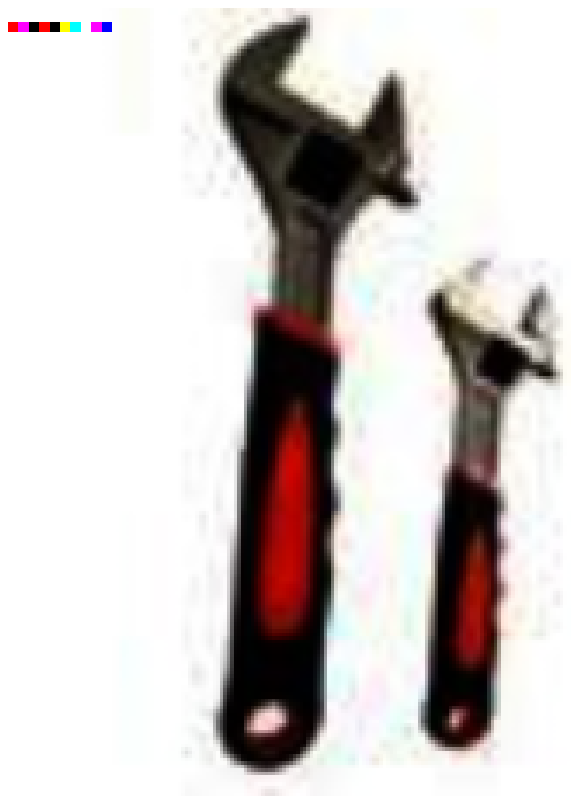}} & 
  axis 28831 chiave inglese regolabile con impugnatura morbida e rullo estremamente scorrevole 200 mm $\dots$ & 
  \fbox{\includegraphics[width=0.215\textwidth]{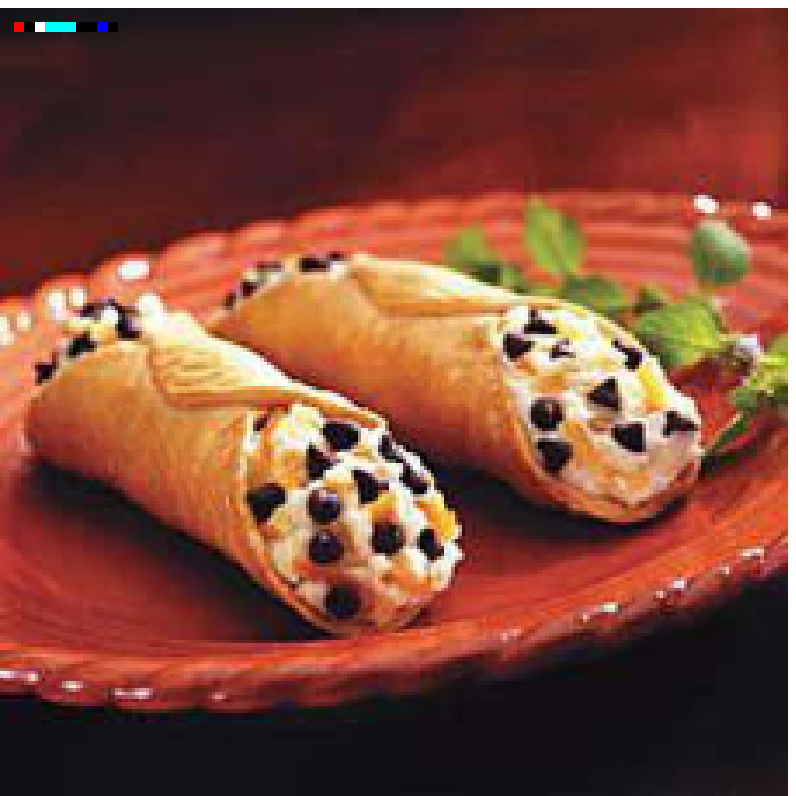}} & Cake Boss Cannoli Cake Ideas and Designs \\ 
  \fbox{\includegraphics[width=0.215\textwidth]{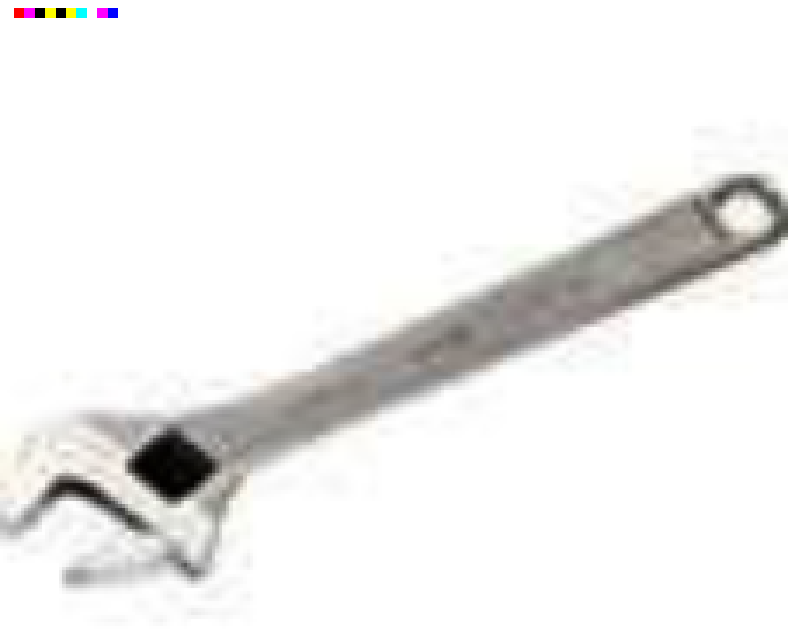}} & 
  sam outillage 54 c10 chiave a rullino cromata 10 lunghezza 255 mm sam & 
  \fbox{\includegraphics[width=0.215\textwidth]{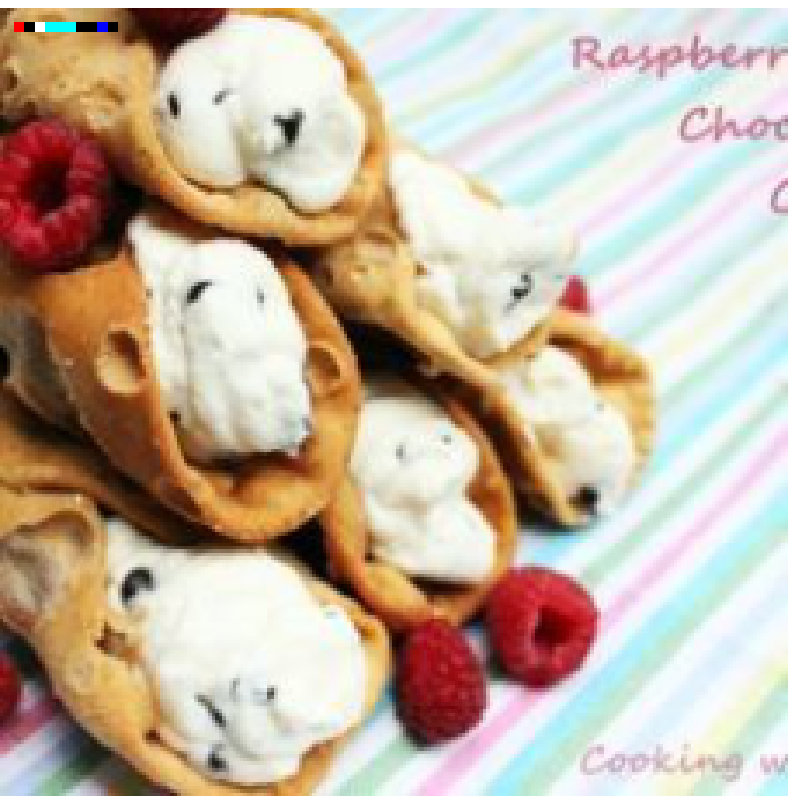}} & Scones* Biscotti* Cannoli on Pinterest \\
  \fbox{\includegraphics[width=0.215\textwidth]{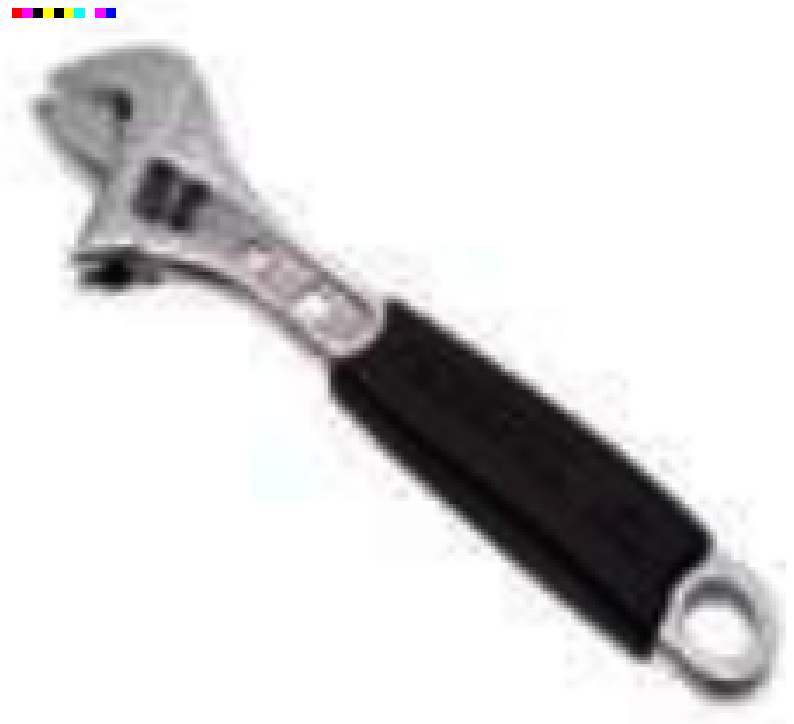}} & 
  faithfull chiave regolabile 150 mm  & 
  \fbox{\includegraphics[width=0.215\textwidth]{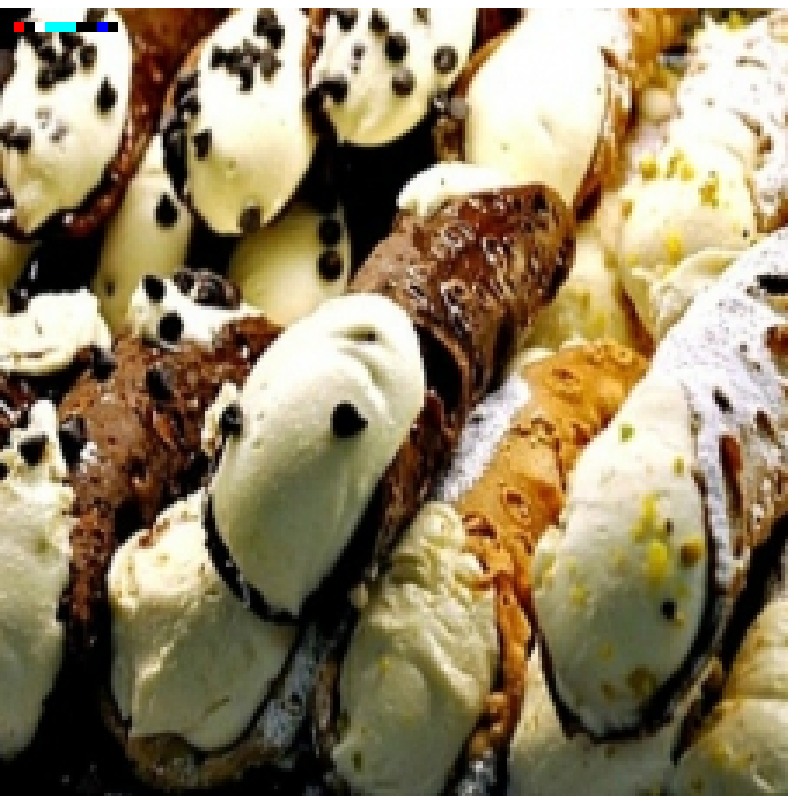}} & Sicilian Cannoli Recipe  The Daily Meal \\
  \end{tabular}
  \caption{
  Each column contains $5$ images and associated text descriptions belonging to a particular class of Ferramenta and UPMC Food-101 datasets.
   Futhermore, each image contains the proposed encoded text.
   Note that the text encodings on each column are similar to each other even if the text and images are different from each other.
  }
  \label{fig:very-similar-encoding}
\end{figure*}

The first experiment consists of extracting only text descriptions from multi-modal datasets, then we train text classification model shown in Figure~\ref{fig:proposed-model}.
Results are shown in first column of Table~\ref{tab:main-results}. 
It is very important to observe how the text encoding extracted is similar to each other when the text description represents similar objects, even when the text information and the images are different from each other (see the example of text encoding in Figure~\ref{fig:very-similar-encoding}).

The second experiment consists of extracting only images from multi-modal datasets, then we train AlexNet~\cite{krizhevsky2012imagenet} and GoogleNet~\cite{Szegedy_2015_CVPR} CNNs from scratch using DIGITS. 
Second and third columns of  Table~\ref{tab:main-results} shows these results.
Images in Ferramenta multi-modal dataset contain objects on a white background, this explains excellent classification results obtained on images alone. 
On the contrary, images in the UPMC Food-101 multi-modal dataset are with complex background and extracted from different contexts, which leads to a low classification performance on images only.

The third experiment consists of employing fused images from multi-modal datasets. 
We train AlexNet~\cite{krizhevsky2012imagenet} and GoogleNet~\cite{Szegedy_2015_CVPR} CNNs from scratch using DIGITS. 
Results in Table~\ref{tab:main-results} indicate that the proposed fusion approach outperforms uni-modal methods. 
Furthermore, the approach is language independent, Ferramenta text descriptions are in Italian.   
Results on UPMC Food-101 clearly indicate benefit of our proposed approach, increasing the classification performance by two folds.
This performance gain is due to leveraging on multi-modal representation learning.

In fourth experiment, we compare our approach with early and late fusion as shown in Table~\ref{tab:early-late-results}.
Experimental setting is inspired from the work~\cite{gallo2017multimodal}.
In particular we use \textit{Logarithmic Opinion Pool}~\cite{Rokach:2010:Ensemble} as a late fusion approach using Random Forest model applied to the 1000 Bag-of-Words while as early fusion we use a Support Vector Machine on the concatenation of Doc2Vec features and 4096 visual features from a trained CNN.
Our proposed approach surpasses standard early and late fusion strategies which further reinforces strength of our approach.

The Figure~\ref{fig:text-vs-multimodal} explores text embedding dimension sizes against $3$ different CNN based architectures i.e. text only, image only and fused image. We see that with lower text-embedding dimension, the fused architecture has an increased performance as compared to the text only architecture. Eventually, both architectures plateau as embedding dimension increases.  
However, the fused image architecture always maintains the upper bound over the other.

\section{Conclusion}
In this work, we proposed a new approach to merge image with their text description so that any CNN architecture can be employed as a multi-modal classification system.
To the best of our knowledge, the proposed approach is the only one that simultaneously exploits text and image casted to a single source, making it possible to use a single classifier.
We obtained promising results and the classification accuracy achieved using our approach is always higher compared to fusion strategies or single modalities.

Another very important contribution of this work concerns the joint representation into the same source of two heterogeneous modalities.
This aspect paves the way to a still open set of problems related to the translation from one modality to another where relationships between modalities are subjective.

\bibliographystyle{IEEEtran}
\bibliography{bib}

\end{document}